\documentclass[journal]{IEEEtran}
\usepackage{amsmath,amsfonts}
\usepackage{algorithmic}
\usepackage{algorithm}
\usepackage{array}
\usepackage{textcomp}
\usepackage{stfloats}
\usepackage{url}
\usepackage{verbatim}
\usepackage{graphicx}
\usepackage{cite}
\usepackage{xcolor}
\usepackage{booktabs}
\usepackage{threeparttable}
\usepackage{multicol}
\usepackage{multirow}
\usepackage[labelformat=simple]{subcaption}
\definecolor{Redtext}{RGB}{139,0,0}
\definecolor{Greentext}{RGB}{0,139,0}
\usepackage[
colorlinks,
linkcolor=red,
anchorcolor=blue,
citecolor=green]{hyperref}
\begin{document}

\title{Balanced Representation Learning for Long-tailed\\Skeleton-based Action Recognition}

\author{Hongda Liu, Yunlong Wang, Min Ren, Junxing Hu, Zhengquan Luo, Guangqi Hou, Zhenan Sun
\thanks{H. Liu and J. Hu are with the School of Artificial Intelligence, University of Chinese Academy of Sciences, Beijing 100049, China, and also with the Center for Research on Intelligent Perception and Computing, State Key Laboratory of Multimodal Artificial Intelligence Systems, Institute of Automation, Chinese Academy of Sciences, Beijing 100190, China (e-mail: hongda.liu@cripac.ia.ac.cn; junxing.hu@cripac.ia.ac.cn).}
\thanks{M. Ren is with the School of Artificial Intelligence, Beijing Normal University, Beijing 100875, China (e-mail: renmin@bnu.edu.cn).}
\thanks{Z. Luo is with the University of Science and Technology of China, Hefei 230027, China, and also with the Center for Research on Intelligent Perception and Computing, State Key Laboratory of Multimodal Artificial Intelligence Systems, Institute of Automation, Chinese Academy of Sciences, Beijing 100190, China (e-mail: zhengquan.luo@cripac.ia.ac.cn).}
\thanks{Y. Wang, G. Hou, and Z. Sun are with the Center for Research on Intelligent Perception and Computing, State Key Laboratory of Multimodal Artificial Intelligence Systems, Institute of Automation, Chinese Academy of Sciences, Beijing 100190, China (e-mail: yunlong.wang@cripac.ia.ac.cn; gqhou@nlpr.ia.ac.cn; znsun@nlpr.ia.ac.cn). Z. Sun is the corresponding author.}
}

\maketitle
\begin{abstract}
Skeleton-based action recognition has recently made significant progress. 
However, data imbalance is still a great challenge in real-world scenarios.
The performance of current action recognition algorithms declines sharply when training data suffers from heavy class imbalance. 
The imbalanced data actually degrades the representations learned by these methods and becomes the bottleneck for action recognition. 
How to learn unbiased representations from imbalanced action data is the key to long-tailed action recognition.
In this paper, we propose a novel balanced representation learning method to address the long-tailed problem in action recognition.
Firstly, a spatial-temporal action exploration strategy is presented to expand the sample space effectively, generating more valuable samples in a rebalanced manner.
Secondly, we design a detached action-aware learning schedule to further mitigate the bias in the representation space. 
The schedule detaches the representation learning of tail classes from training and proposes an action-aware loss to impose more effective constraints.
Additionally, a skip-modal representation is proposed to provide complementary structural information. 
The proposed method is validated on four skeleton datasets, NTU RGB+D 60, NTU RGB+D 120, NW-UCLA, and Kinetics.
It not only achieves consistently large improvement compared to the state-of-the-art (SOTA) methods, but also demonstrates a superior generalization capacity through extensive experiments.
Our code is available at \href{https://github.com/firework8/BRL}{https://github.com/firework8/BRL}.
\end{abstract}


\section{Introduction}
\IEEEPARstart{H}{uman} action recognition has witnessed many advances in recent years. 
It is of vital importance in many applications, such as video analysis and behavior understanding~\cite{shahroudy2016ntu,liu2019ntu,sun2022human}. 
Multiple modalities, such as video, depth images, optical flows~\cite{simonyan2014two}, and body skeletons, have been employed for action recognition.  
In particular, skeleton-based human action recognition has attracted much attention mainly due to the compact motion-related structural information and the noise-free background~\cite{wang2014cross}.

In general, skeleton data can be efficiently obtained by depth sensors~\cite{zhang2012microsoft} and advanced human pose estimation algorithms~\cite{cao2017realtime}. 
Researchers construct some large, representative datasets~\cite{wang2014cross,shahroudy2016ntu,liu2019ntu,kay2017kinetics} by collecting and processing skeleton data. These datasets are usually resampled and relatively balanced among action categories for research purposes. Nevertheless, these academic datasets are not biased in favor of real-world circumstances, viz., realistic action data are imbalanced.
As real-world action categories become more diverse and fine-grained, there will be a long-tailed situation where frequent action classes have massive samples but rarely-occurring action classes are associated with only a few samples. 
It is not surprising that current recognition methods~\cite{yan2018spatial,shi2019two,liu2020disentangling,duan2022pyskl} cannot deal with the long-tailed imbalance situation well. We experimentally observed that the performance of different methods deteriorates significantly when directly tested on long-tailed simulated action datasets. Further, their deteriorated performances are close to each other. Without deliberately altering other factors, this indicates a strong connection that the representations learned by these methods are generally degraded due to the data imbalance between head and tail classes.

From the perspective of representation learning, imbalanced data brings two problems. 
First, imbalanced data directly leads to a skewed sample space, which is not well accommodated by current recognition methods.
The skewed sample space will cause recognition models to have a bias towards the dominant class and perform poorly on the tail class. 
Due to the inherent differences between skeletal and image data, current approaches~\cite{shen2016relay,mullick2019generative,park2022majority} that focus on long-tailed classification are not well suited to long-tailed action recognition. 
For action recognition, the spatial and temporal distinctive information of skeletal data is important. 
Rather, existing methods do not take such structural information into account well, ultimately degrading the quality of the samples.
Increasing the sample quantity redundantly and repetitively using these approaches may lead to information overlap, and the marginal benefits from the data are diminishing to the model~\cite{buda2018systematic,cui2019class}. 
This fact suggests that how to generate more valuable skeleton samples tailored for action recognition is critical to complementing the skewed sample space.

Second, imbalanced data results in representation bias.
Due to the imbalanced distribution, the recognition model inclines to encode more discriminate factors of head classes, which distorts the representation space of tail classes.
The representation bias would then confuse the classifier, resulting in misclassification.
Many current methods~\cite{zhang2021deep,wu2020distribution,tan2020equalization,li2022long} adjust the weights of different classes to mitigate bias.
However, these re-weighting methods improve the performance of tail classes at the cost of head-class performance, which traps in a performance seesaw.
Moreover, these methods do not consider generic knowledge between different action classes to compensate for imbalanced bias.
For long-tailed action recognition, consensus knowledge in head and tail action classes are both important, which demands a balance in the objective of representation learning.
Therefore, how to learn balanced representations needs to be seriously considered for mitigating the representation bias in long-tailed action recognition.

\begin{figure*}[t]
\vspace{-0.5cm}
\centering
\includegraphics[width=\linewidth]{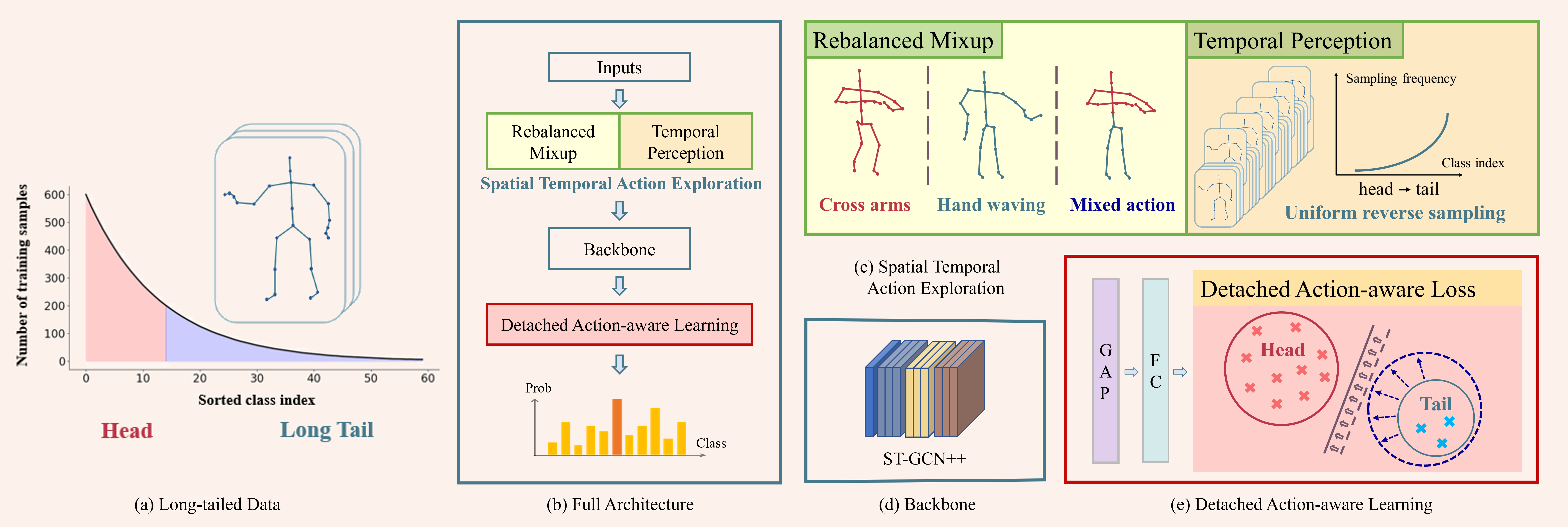}
\setlength{\abovecaptionskip}{0pt}
\setlength{\belowcaptionskip}{0pt}
\caption{ 
\textbf{Architecture Overview}. 
(a) Long-tailed action data. The head class and tail class of a long-tailed action dataset (NTU RGB+D-LT in this sample) have drastically different numbers of samples. (b) An overview of the proposed architecture. (c) Spatial-temporal action exploration strategy to generate more valuable skeletal data. (d) ST-GCN++ is adopted as the backbone. (e) Detached action-aware learning is proposed to mitigate representation bias in long-tailed action recognition.}
\label{fig:figure1}
\vspace{-0.1cm}
\end{figure*}

Faced with the above two dilemmas, a balanced representation learning method is proposed for long-tailed skeleton-based action recognition. 
To generate more valuable samples, a novel spatial-temporal action exploration strategy is presented. 
The proposed strategy contains two modules, i.e., rebalanced partial mixup and temporal reverse perception. 
On the one hand, rebalanced partial mixup is designed to strengthen spatial structure information of skeletal action data. 
This module mixes the skeleton part of different bodies in a rebalanced manner to generate new samples. 
The rebalanced manner means using different mixing factors in the sample space and label space.
With an adaptive label factor designed to rebalance the label space, the method can assign higher weights to minority categories and alleviate the effects of imbalanced action data.
On the other hand, temporal reverse perception is presented to enhance temporal information.
This module uniformly divides the original sequence into splits with equal lengths and randomly takes more samples for the tail classes, which drives the sampler to take excessive enhancements for the tail classes.
In addition, a skip-modal representation of the joints is adopted for multi-modal fusion.
After mining different modal steams, the model can introduce additional structural information for tail classes.
By coupling these components, spatial-temporal action exploration provides more valuable samples in spatial and temporal domains collaboratively, thereby complementing the skewed sample space.

To mitigate representation bias, we propose a detached action-aware learning schedule.
The recognition model initially learns general knowledge of skeleton sequences through a vanilla cross-entropy loss. 
Subsequently, the action-aware loss is introduced to drive the model to focus on specific patterns of tail classes.
After calculating sample effectiveness and category relative frequencies, an action-aware balanced term is added to the loss function in a model-agnostic manner, resulting in more effective constraints imposed on different classes.
In this way, we detach specific patterns of tail classes from the learning of general knowledge and enhance them effectively.
Fig.~\ref{fig:figure1} illustrates our method, and extensive experimental results demonstrate the schedule mitigates existential bias in the representation space and provides a significant performance boost.

The main contributions of this paper can be summarized as follows:
\begin{itemize}
\item 
The bottlenecks of long-tailed skeleton-based action recognition are explored. 
It is investigated that generating valuable samples and mitigating representation bias are key factors determining the performance of recognition models. Based on the observations, a balanced representation learning method is proposed.

\item 
A spatial-temporal action exploration strategy is presented, which generates more valuable skeletal samples.
In addition, a detached action-aware learning schedule is proposed to mitigate the representation bias learnt from imbalanced data. 

\item 
Extensive experiments and ablation studies prove the effectiveness of the proposed method.
Overall, the proposed method can be easily integrated into various action recognition frameworks at the cost of a slight computational overhead.
\end{itemize}

The remainder of this paper is organized as follows: 
Section~\ref{paper_2} presents a brief literature review of the related work. 
The proposed method is described in detail in Section~\ref{paper_3}.
The configurations and results of extensive experiments are presented in Section~\ref{paper_4}.
Finally, the conclusions of this paper are summarized in Section~\ref{paper_5}.

\section{Related Work}\label{paper_2}

\subsection{Skeleton-based Action Recognition}

In the early stages, traditional skeleton-based action recognition methods mainly focused on designing hand-crafted features~\cite{vemulapalli2014human}.
The hand-crafted features are straightforward and intuitive, but weak in representing the semantic connectivity information of body parts.
With the development of deep learning, convolutional neural networks (CNNs) and recurrent neural networks (RNNs) are applied to action recognition~\cite{tran2015learning,du2015hierarchical,wang2017modeling}. 
Subsequent applications of long short-term memory (LSTM) and attention mechanisms further exploit the spatial and temporal dynamics information~\cite{song2017end,si2018skeleton,si2019attention,zhang2022twinnet}.
However, these methods are still limited in the exploitation of the structural topology of the joints. Considering the inherent graph structure of the skeleton point sequence, graph-based methods can be naturally utilized to deal with the skeleton graph~\cite{song2020richly,zhang2020semantics,song2022constructing,ren2022towards,duan2022revisiting}.
By constructing spatial-temporal graphs, a series of graph-based approaches have been achieved with promising performances, which indicates that graph networks successfully capture the most informative features for various action classes.

Since then, many ingenious graph networks are proposed consecutively. 
Yan \textit{et al.}~\cite{yan2018spatial} firstly apply graph-based neural networks for skeleton-based action recognition, and propose ST-GCN, which combines spatial graph convolutions and temporal dynamics convolutions for spatiotemporal modeling.
Upon the baseline, 2s-AGCN~\cite{shi2019two} proposes an adaptive graph convolutional network, which adaptively models correlation between two joints. 
The two-stream ensemble with skeleton bone is also adopted to boost recognition performance.
Subsequently, Cheng \textit{et al.}~\cite{cheng2020skeleton} present shift graph operations and lightweight point-wise convolutions to improve the efficiency of the model.
A concurrent work by Liu \textit{et al.}~\cite{liu2020disentangling} proposes a multi-scale aggregation scheme and directly models spatial-temporal dependencies.
Correspondingly, CTR-GCN~\cite{chen2021channel} explores topology-non-shared graph convolutions and proposes a channel-wise topology refinement graph convolution.
InfoGCN~\cite{chi2022infogcn} also models context-dependent intrinsic topology, but it further leverages an information-theoretic objective and multiple modalities to better represent latent information.
Recently FR Head~\cite{zhou2023learning} propose a discriminative feature refinement module to improve the performance of ambiguous actions.
Although these methods achieve encouraging performance for skeleton-based action recognition, how to design the networks to address skeleton-based action recognition is still an under-explored important question.
Meanwhile, there is much room for improvement, especially from the aspect of multi-modal fusion.
The way to fuse features from different modalities~\cite{shi2019two,chi2022infogcn} needs to be carefully investigated. 
In addition, some recent work~\cite{guo2022contrastive,moliner2022bootstrapped} focus on self-supervised action recognition and other interesting aspects.

However, as far as we know, research on long-tailed skeleton-based action recognition is still blank. 
It is believed that our work can shed light on the weaknesses of action recognition when facing long-tailed data distributions, and inspire the following work to further investigate the intrinsic nature of action recognition.

\subsection{Long-tailed Learning}

Long-tailed dilemmas have attracted increasing attention due to the prevalence of imbalanced data in real-world applications~\cite{huang2016learning,cui2019class}. 
The methods on long-tailed imbalanced data can be simply divided into two regimes: re-sampling and cost-sensitive learning.
Re-sampling methods~\cite{park2022majority} focus on under-sampling the head category and over-sampling the tail category, which aim to modify the training distributions to decrease the level of imbalance.
Cost-sensitive learning can be traced back to importance sampling in statistics, which assigns weights to samples to match a given data distribution~\cite{li2022key}.
Through cost-sensitive loss function, re-weighting methods direct the network to allocate more attention to the samples in tail classes than head classes.
Recent studies~\cite{zhang2021deep} make progress in the following three directions: information augmentation, class re-balancing, and module improvement.
Specifically, data augmentation~\cite{mullick2019generative,chou2020remix,park2022majority}, transfer learning~\cite{liu2019large}, logit adjustment~\cite{cao2019learning,ren2020balanced,alexandridis2022long}, decoupled training~\cite{kang2019decoupling}, ensemble learning~\cite{zhou2020bbn}, and semi-supervised learning~\cite{yang2020rethinking} are further combined with classical long-tailed learning methods.

\subsection{Long-tailed Skeleton-based Action Recognition}

Current long-tailed learning work mainly concentrate on image classification~\cite{liu2019large,jia2022multi}, detection~\cite{tan2020equalization}, instance segmentation~\cite{zhang2021deep}, multi-label classification~\cite{wu2020distribution}, and video classification~\cite{zhang2021videolt}. While extensive studies have been done for the long-tailed tasks above, little effort has been made for long-tailed action recognition. Compared with the image or video frames, the skeleton sequence is more generalized and condensed.
The topological connections of joint points in the skeleton contain extensive spatial-temporal structural information.
Due to the specificity of skeleton data, current algorithms are not well suited for skeleton-based action recognition.
Although some existing works~\cite{zhang2017mixup,xu2022topology,chen2022contrastive} attempt to adopt structural information, they are unaware of addressing the imbalance of the action data.
Additional re-weighting approaches propose cost-sensitive factors such as class prediction hardness~\cite{lin2017focal} and effective number~\cite{cui2019class} to impose constraints. But for skeleton graph networks, imposing constraints on the head classes throughout training will affect the learning of discriminative action representations and lead to optimization difficulties.

In contrast to the literature, a balanced representation learning method is proposed in this paper to specifically resolve the problem of long-tailed skeleton-based action recognition. 
Combining the philosophy of long-tailed learning, the proposed method provides a new perspective for understanding long-tailed action data especially by enhancing the sample space and mitigating representation bias.

\section{Methodology}\label{paper_3}

\begin{figure*}[ht]
\vspace{-0.5cm}
\centering
\includegraphics[width=\linewidth]{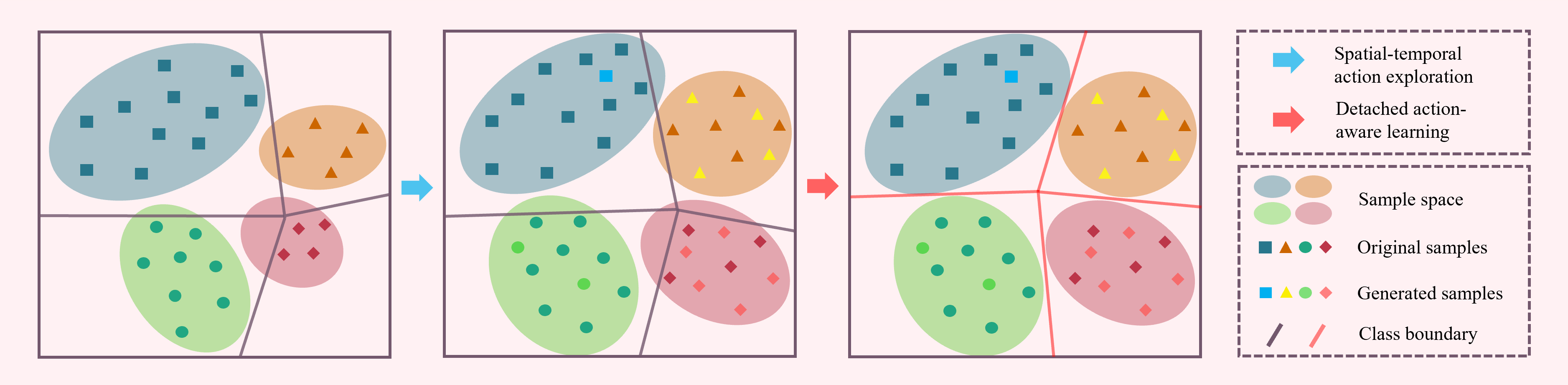}
\setlength{\abovecaptionskip}{0pt}
\setlength{\belowcaptionskip}{0pt}
\caption{ 
An illustration of the proposed method. 
(Best viewed in color.) 
First, the spatial-temporal action exploration strategy is presented to expand the corresponding sample space effectively.
Besides, due to skewed data distribution, the original decision boundary is compromised.
The detached action-aware learning schedule can delicately adjust decision boundaries to further mitigate representation bias.
Through the unification process, the proposed method can learn the balanced representations from long-tailed action data.
}
\label{fig:figure2}
\vspace{-0.1cm}
\end{figure*}

The details of the proposed balanced representation learning method will be elaborated in this section and an illustration of our method is visualized in Fig.~\ref{fig:figure2}.

\subsection{Data Pre-processing}\label{method_a}

For skeletal data pre-processing, some current approaches generally adopt the augmentation solution provided by~\cite{zhang2020semantics}.
Effective augmentation for action samples can assist network learning.
For representation learning of long-tailed data, the choice of data augmentation is even more critical. We absorb the previous pre-processing approach and design a more efficient augmentation scheme tailored for long-tailed action data.

The normal skeleton augmentations basically include six spatial augmentations, i.e. \emph{Flip}, \emph{Rotate}, \emph{Shear}, \emph{Scaling}, \emph{Mask}, \emph{Part Drop}, and four temporal augmentations, i.e. \emph{Temporal Crop}, \emph{Temporal Flip}, \emph{Temporal Shift}, and \emph{Sampling}. 
Filters can also be used for spatial and temporal augmentations, such as \emph{Gaussian Noise} and \emph{Gaussian Blur}. 
Furthermore, the augmentation of skeleton sequences is different from normal images and videos, which should care more about the spatial and temporal distinctive information. Thus, such structural information needs to be leveraged and further enhanced when augmenting long-tailed skeletal data.
As the combination of augmentations is complicated, we make it a priority to select augmentation combinations that can introduce more novel movement information.
Extensive experiments are conducted to test the effect of different strategies.

It is experimentally verified that \emph{Flip}, \emph{Rotate}, and \emph{Scaling} in the spatial augmentations and \emph{Sampling} in the temporal augmentations are more powerful for long-tailed action data. 
The four augmentation strategies are integrated with the current pre-processing method.
The method is used to finally get diverse skeleton sequences.

\subsection{Spatial-Temporal Action Exploration}\label{method_b}

This section details the rebalanced partial mixup and temporal reverse perception.

\subsubsection{Rebalanced Partial Mixup}
Mixup~\cite{zhang2017mixup} was proposed as a regularization technique for improving the generalization of train data. 
The method constructs virtual training examples and extends the training distribution by linear interpolations of feature vectors. 
It works by mixing two images and their labels linearly as:
\begin{equation}\label{eq:mixup}
\widetilde{x} = \lambda x_{i} + (1 - \lambda)x_{j}\\
\end{equation}
\begin{equation}
\widetilde{y} = \lambda y_{i} + (1 - \lambda)y_{j}
\end{equation}
where $(x_{i}, y_{i})$ and $(x_{j}, y_{j})$ are two examples drawn at random from the training data, and $\lambda \in [0, 1]$.

For skeleton-based action recognition, spatial structure information is necessary to understand the action.
Moreover, the same skeleton parts have distinctive significance in different actions.
Hence, for generating meaningful samples to attenuate the skew in sample space, the proposed rebalanced partial mixup considers mixing the skeleton part of different bodies in a rebalanced manner.

The rebalanced manner means using two respective mixing factors to mix up in the sample space and label space.
This is because the previous single mixing factors simply extend the training sample space, but do not further address the imbalance of the action data.
The proposed strategy thereby turns to disentangle the mixing process of sample space and label space.
An adaptive label factor is designed to provide an effective regularization focusing on tail classes.
Through rebalancing the label space, the strategy can assign a higher weight to the minority class and alleviate the effects of imbalanced action data.

The formulation of the rebalanced partial mixup is defined as follows:
\begin{eqnarray}\label{eq:rebalanced_mixup}
\widetilde{x} &=& Concat (\lambda_{x}x_{i}, (1 - \lambda_{x})x_{j})\\
\widetilde{y} &=& \lambda_{y} y_{i} + (1 - \lambda_{y}) y_{j}
\end{eqnarray}
where $x_{i}, x_{j}$ are the joint indices of the body. 
$\lambda_{x}$ means the ratio of selected joints for the whole body. 
The exact form of $\lambda_{y}$ is defined according to the following formula:
\begin{equation}
\label{equ:lamada_y}
\lambda_{y}=\left\{
\begin{array}{lcl}
0, & & \dfrac{n_{i}}{n_{j}} > k\\
1, & & \dfrac{n_{i}}{n_{j}} \leq \dfrac{1}{k} \\
\lambda_{x}, & & otherwise
\end{array} 
\right.
\end{equation}

\begin{figure}[t]
\centering
\includegraphics[width=\linewidth]{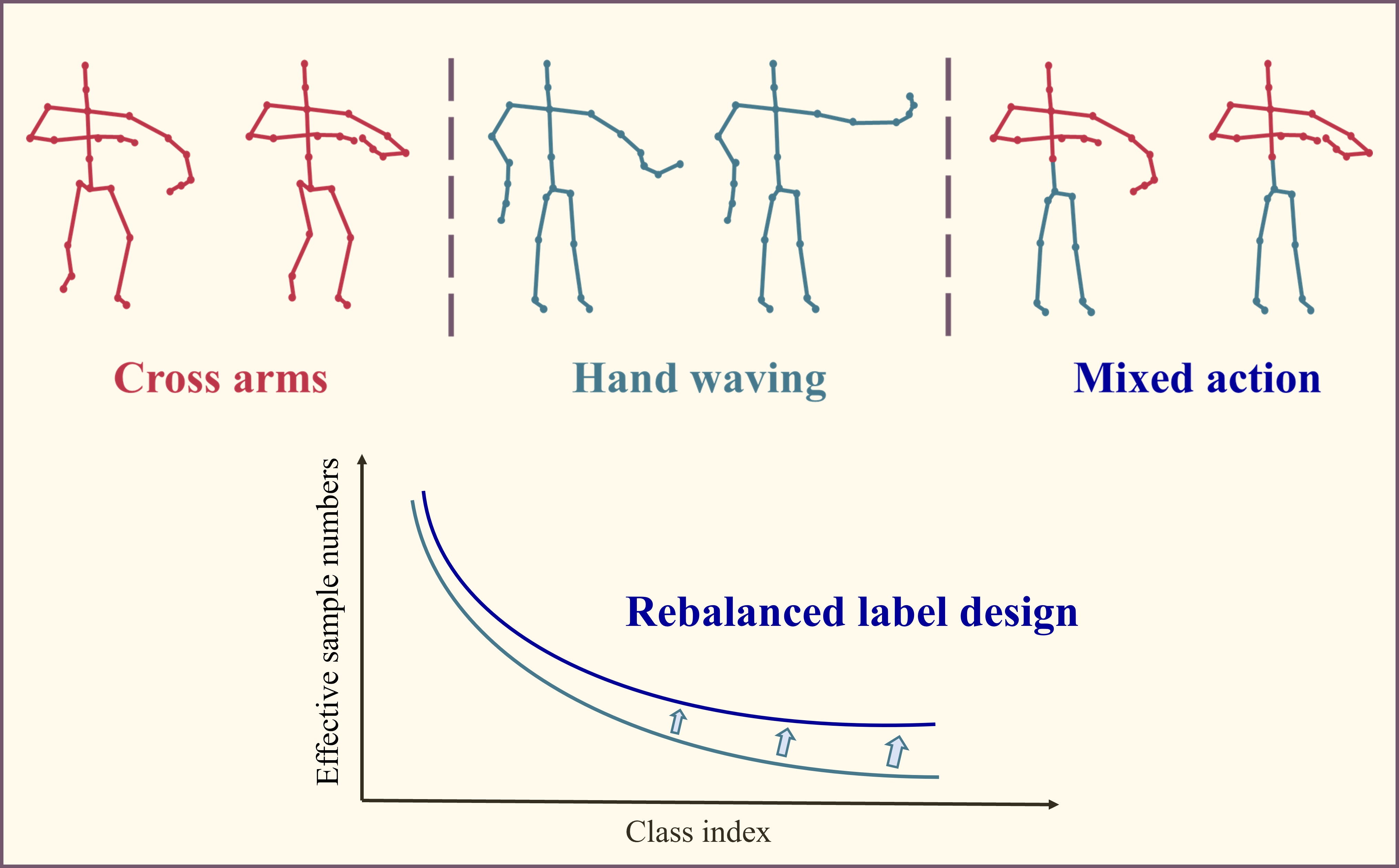}
\setlength{\abovecaptionskip}{0pt}
\setlength{\belowcaptionskip}{0pt}
\caption{
Rebalanced part mixup combines parts of the body skeleton to generate more valuable skeleton data in a rebalanced manner.
With the rebalanced label design, the strategy can alleviate the effects of imbalanced data, resulting in a more balanced distribution of action data.
}
\label{fig:figure3}
\vspace{-0.1cm}
\end{figure}

Here $n_{i}$ and $n_{j}$ denote the number of samples in the corresponding action classes that sample $i$ and sample $j$ belong to. 
$k$ is a hyper-parameter, which indicates the degree of imbalance between the two categories.
Specifically, if $n_{i}$ is more than $k$ times larger than $n_{j}$, $\lambda_{y}$ is set to $0$, less than $1/k$, $\lambda_{y}$ is set to $1$, otherwise, $\lambda_{y}$ is equal to $\lambda_{x}$. 
Different parts of the body are chosen to formulate new samples, and the corresponding rate is adjusted accordingly. 
The radio for selecting random samples is set to $1/16$, which means choosing one sample per $16$ samples.
We adopt combining the upper body and the lower body directly as shown in Fig.~\ref{fig:figure3} and apply the rebalanced label design.

Actually, the key to the proposed method is generating meaningful samples in a rebalanced manner. 
The rebalanced partial mixup can effectively aggregate the spatial structure information of long-tailed action samples, expanding the space of skeleton samples.
And the rebalanced label design rebalances the label space and alleviates the effects of imbalanced data.
The performance can be further improved by the effective regularization technique.
In this way, the strategy generates more meaningful samples.

\subsubsection{Temporal Reverse Perception}
Temporal distinctive information is also important for long-tailed action data.
For the oversampling of imbalanced skeleton data, the sub-string sampling with bilinear interpolation does not make full use of the temporal information of the skeleton sequences.
Hence, temporal reverse perception is proposed to help the model better understand temporal information in skeletal data.

First, the original input sequence is divided uniformly into some splits with equal lengths and then one frame is randomly selected from each split. 
Through the approach, the sampler can generate more distinctive skeleton sequences with similar temporal attributes. 
And the temporal integrity of the skeleton sequences is well-preserved after multiple random samplings.
In addition, the set of tested skeleton sequence lengths is $\{300, 100, 64, 50, 32, 16\}$.
It is experimentally verified that $64$ is the best for long-tailed skeleton data.
The uniform sampler effectively maintains sample variability, especially when combined with the rebalanced partial mixup method.

Furthermore, due to the limited number of samples in the tail classes, oversampling can lead to information redundancy.
This redundancy of temporal information reduces the benefit of generating more data, which influences representation learning.
The proposed temporal reverse perception can drive the sampler to take special enhancements for the tail classes. 
Specifically, the reverse sampler adaptively selects the sampling frequency according to the actual action data distribution, which can further reduce the invalid overlap and maintain meaningful temporal information.

Briefly, the temporal reverse perception is composed of the uniform sampler and the reverse sampler.
The approach is to sample frames uniformly from the original sequences and take more samples at random for the tail classes.
Combined with rebalanced partial mixup and temporal reverse perception, the proposed spatial-temporal action exploration strategy can provide more valuable data for the network to learn better representation.

\subsection{Detached Action-Aware Learning Schedule}\label{method_c}

To mitigate representation bias, the detached action-aware learning schedule is proposed, which contains a detached training strategy and an action-aware loss.

The class re-weighting approach typically provides a way to assign appropriate weights to samples. 
However, for long-tailed action recognition, it is not that easy as the performance will gradually turn worse when using the re-weighting loss for the whole training process, especially in the head classes. 
This indicates that imposing constraints on the head classes throughout training actually affects the learning of general knowledge.

Therefore, we present a detached training strategy. 
It detaches specific patterns of tail classes from the learning of general knowledge, enabling the model to first learn generic action representations from majority classes. Afterwards, it switches to perceive specific patterns of tail classes.
Specifically, the model first learns generic knowledge of action recognition through normal CE loss.
After several epochs, a transition from CE loss to an action-aware loss drives the model to learn special action patterns from tail data and achieve a more comprehensive representation space.
The strategy avoids neglecting the existence of tail classes in the early training process, or over-suppressing head classes in the latter re-weighting strategy, resulting in more effective constraints being imposed on different classes during training.

For an input action sample $x$ with label $y \in \{1,..., C\}$, suppose that the predicted output from the model for all classes are $ {\rm \textbf{z}} =  [z_{1},...,z_{C}]^\top$, where $C$ is the total number of classes.
Given a sample with class label $y$, the Softmax loss function for this sample is written as:
\begin{equation}
\label{equ:e4}
\mathcal{L}({\rm z}, y)=-\log p_{y}, {\rm with}\  p_{y}=\dfrac{\exp(z_{y})}{\sum_{j=1}^C \exp(z_{j})}
\end{equation}
where the subscript $y$ denotes the target class. Here, $z_{y}$ indicates the target logit, and accordingly, $z_{j}$ denotes the predicted logit of class $j$. Therefore, $p_{y}$ denotes the estimated class probability.

To reflect the effectiveness of different action samples adaptively, we further propose the action-aware term, which is defined as:
\begin{equation}
\beta_{y} = \lambda \times \dfrac{n_{y}-n_{min}}{n_{max}-n_{min}} + \upsilon
\end{equation}
where $n_{y}$ is the number of samples for class $y$, $n_{max}$ and $n_{min}$ denote the maximum and minimum sample number of all classes in the corresponding data.
$\lambda$ and $\upsilon$ are hyperparameters for range scaling, which are set according to the imbalance ratio of action data.
Through calculating category relative frequencies as above, $\beta_{y}$ will be restricted to the fixed range $[\upsilon, \lambda+\upsilon]$.
With scaling restriction, the term can make the weights of action classes more distinctive.

The weighting factor $\gamma$ will be formulated as:
\begin{equation}
\gamma = \dfrac{1}{\sum_{i=1}^{n} \beta^{i-1}}=\dfrac{\beta-1}{{\sum_{i=1}^{n} \beta^{i-1}} \times (\beta-1)}=\dfrac{\beta-1}{\beta^{n}-1}
\end{equation}
where $\beta$ is the action-aware term. 
Here, consider the case where the model has previously sampled $n-1$ examples and needs to sample $n$-th sample. 
$\sum_{i=1}^{n} \beta^{i-1}$ means the $i$-th sample contributes $\beta^{i-1}$ to the effective number $\sum_{i=1}^{n} \beta^{i-1}$. 
The action-aware term $\beta$ is calculated by the relative probability factor and scaling hyperparameters, instead of the number of samples $N$.

Due to $\beta \in (0,1)$, for the intuitive calculation and consistency with the previous form, $\gamma$ is calculated as:
\begin{equation}
\gamma =\dfrac{1-\beta}{1-\beta^{n}}
\end{equation}

Suppose class $y$ has $n_{y}$ training samples, the action-aware loss can be expressed as follows:
\begin{equation}
\label{equ:e9}
\mathcal{L}_{action-aware}({\rm z}, y)=- \dfrac{1-\beta_{y}}{1-\beta_{y}^{n_{y}}} \log \dfrac{\exp(z_{y})}{\sum_{j=1}^C \exp(z_{j})}
\end{equation}

With the proposed action-aware loss, the model will pay more attention to the tail classes and avoid excessive training of the tail classes.
Further, the classifier can better distinguish the samples from tail action classes that are prone to be confused.

After combining the detached training strategy with the action-aware loss, the method can enable all samples to participate in the training more efficiently in order to converge quickly.
For learning the detached tail representations, the method can better understand specific long-tailed patterns, while maintaining head performance.
In general, the proposed detached action-aware learning schedule mitigates the representation bias effectively.
The overall training procedure is summarized in Algorithm~\ref{alg:algorithm}.

\begin{algorithm}[htbp]
\caption{Detached Action-Aware Learning Schedule}
\label{alg:algorithm}
\textbf{Require}: Training dataset $\emph{D}=\{(x_{i}, y_{i})\}_{i=1}^{n} $; \\
\mbox{} \qquad  \quad $\:\,$ A parameterized model $f_{\theta}$. \\
\textbf{Output}: Updated model $f_{\theta}$.
\begin{algorithmic}[1]
\STATE Initialize the model parameters $\theta $ randomly;
\FOR{ $t=1$ to $T_{0}$}
\STATE Sample batch samples $\mathcal{B}$ with batch size of $m$:\\
$\mathcal{B}\xleftarrow[]{} $SampleBatch$(D,m)$;
\STATE Calculate the loss by Eq. (\ref{equ:e4}):\\
{\color{Greentext} $\mathcal{L}({f}_{\theta})\xleftarrow[]{} \dfrac{1}{m}\sum_{(x,y)\in\mathcal{B}}\mathcal{L}_{\rm CE}((x,y);{f}_{\theta})$}
\STATE ${f}_{\theta}\xleftarrow[]{}{f}_{\theta}-\alpha \nabla \mathcal{L}({f}_{\theta});$
\ENDFOR
\FOR{ $t=T_{0}$ to $T$}
\STATE Sample batch samples $\mathcal{B}$ with batch size of $m$:\\
$\mathcal{B}\xleftarrow[]{} $SampleBatch$(D,m);$
\STATE Calculate the loss by Eq. (\ref{equ:e9}):\\
{\color{Redtext} $\mathcal{L}({f}_{\theta})\xleftarrow[]{} \dfrac{1}{m}\sum_{(x,y)\in\mathcal{B}}\mathcal{L}_{\rm action-aware}((x,y);{f}_{\theta})$}
\STATE ${f}_{\theta}\xleftarrow[]{}{f}_{\theta}-\alpha \nabla \mathcal{L}({f}_{\theta}).$
\ENDFOR
\end{algorithmic}
\end{algorithm}
\vspace{-0.1cm}

\subsection{Ensemble with Multi-Modal Representation}\label{method_d}

In contrast to normal action recognition under the balanced data setting, multi-modal fusion is more helpful for long-tailed data. 
The cause of the phenomenon is further analyzed, that is, different modalities provide a natural data augmentation for long-tailed data. 
Hence, we extend the spatial structure and define a new modal representation named skip-modal representation.

The original joint modality corresponds to the coordinate information of the skeleton joint points in space.
Subsequent bone modality is represented as the vectors pointing to the target joint from the source joint.
The proposed skip-modal representation is to skip one joint point on top of the bone modality.
It can obtain more condensed spatial relationship information while maintaining directional information.
The representation can explore the deeper connections between joints to analyze the actions. 
Other modalities of skipping more joint steps are also investigated, but the performances are not as good as the skip-one-joint step.
Hence, we mainly use the skip-one-joint-step representation as the adopted skip-modal representation.

After training the model with each modal representation, the results of different modalities are ensembled during inference.
The multi-modal fusion actually provides complementary information for obtaining more discriminative representations, leading the model to understand long-tailed action data more comprehensively.

\section{Experiments}\label{paper_4}

In this section, we evaluate the performance of the proposed method on four skeleton-based action recognition datasets.
Ablation studies are also performed to validate the effectiveness of each component in the model.
Finally, analysis of experimental results and visualizations are reported to further demonstrate the efficacy of the proposed method.

\subsection{Datasets}

\subsubsection{NTU RGB+D 60}
This indoor captured dataset~\cite{shahroudy2016ntu} is a large-scale human action recognition dataset containing 56,880 skeleton action sequences of 60 action classes.
The dataset recommends two benchmarks: 
(1) \textbf{cross-subject (X-sub)}: training data comes from 20 subjects, and testing data comes from the other 20 subjects. 
(2) \textbf{cross-view (X-view)}: training data comes from camera views 2 and 3, and testing data comes from camera view 1.

\subsubsection{NTU RGB+D 120}
This is currently the largest indoor action recognition dataset~\cite{liu2019ntu}, which is an extended version of NTU RGB+D with 60 more action classes.
The dataset contains 114,480 videos and consists of 120 classes.
Similarly, the recommended two settings are suggested: 
(1) \textbf{cross-subject (X-sub)}: training data comes from 53 subjects, and testing data comes from the other 53 subjects. 
(2) \textbf{cross-setup (X-set)}: training data comes from 16 even setup IDs,  and testing data comes from 16 odd setup IDs.

\subsubsection{Northwestern-UCLA}
Northwestern-UCLA~\cite{wang2014cross} is captured by using three Kinect cameras.
It contains 1494 samples covering 10 action categories, and each action category is performed by 10 subjects.
We adopt the same evaluation protocol in~\cite{wang2014cross}: training data comes from the first two cameras, and testing data comes from the other camera.

\subsubsection{Kinetics Skeleton 400}
The dataset~\cite{kay2017kinetics} is adapted from the Kinetics 400 video dataset using the OpenPose toolbox in 2D keypoint modality. 
It contains 240,436 training and 19,796 evaluation skeleton clips over 400 classes, where each skeleton graph contains 18 body joints.
At each time step, two people are selected for multi-person clips based on the average joint confidence.
Regarding the experiments using 2D estimated skeleton, we use publicly available HRNet skeletons provided by PoseConv3D~\cite{duan2022revisiting} for fair comparison.
Typically, Top-1 classification accuracy is used in the evaluation protocol.

\subsubsection{Long-tailed datasets}

To validate the effectiveness of our method for long-tailed skeleton data, we construct the long-tailed datasets based on NTU 60, NTU 120, and Northwestern-UCLA.
As stated before, Kinetics 400 is composed of many unique short videos. 
The statistics of samples versus classes indicate that the dataset itself actually exhibits a long-tailed data distribution.
Therefore, we also choose Kinetics Skeleton 400 for the experiments.

\begin{figure}[t]
\centering
\includegraphics[width=\linewidth]{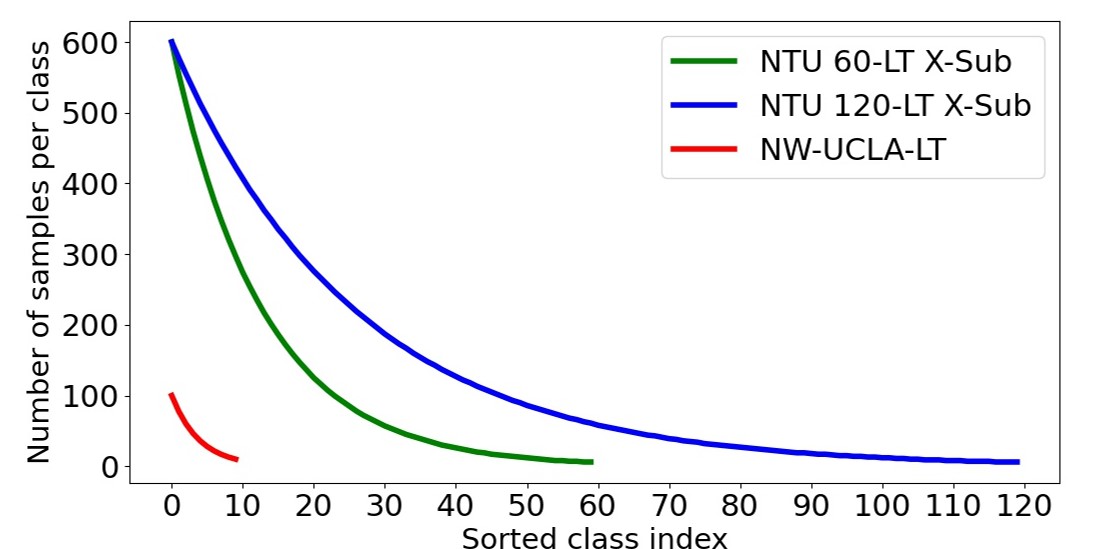}
\setlength{\abovecaptionskip}{0pt}
\setlength{\belowcaptionskip}{0pt}
\caption{
The number of training samples per class in the constructed long-tailed action datasets.
}
\label{fig:figure4}
\vspace{-0.1cm}
\end{figure}

For constructing the long-tailed version of the above datasets, we follow the general setting of long-tailed learning, i.e., truncating a subset with the Pareto distribution from the balanced version~\cite{buda2018systematic}.
Specifically, to create the imbalanced version, the number of training examples per class is reduced, and the validation set is kept unchanged~\cite{cao2019learning}. 
Imbalance ratio $\beta $ is utilized to denote the ratio between sample sizes of the most frequent and least frequent class, i.e., $\beta={\rm max}_{i}\{{n}_{i}\} / {\rm min}_{i}\{{n}_{i}\} $. 
The whole long-tailed imbalance follows an exponential decay in sample sizes across different classes. 
For NTU 60, the maximum number of samples per class under cross-subject and cross-view settings is set as 600. The imbalance ratio is fixed at 100.
As for NTU 120, we adjust the maximum number of samples per class to 600 for the cross-subject setting and 400 for the cross-view setting, taking into account the change in the number of categories and samples. Additionally, we maintain the imbalance ratio at 100.
Regarding Northwestern-UCLA, the maximum number of samples per class is set as 100. 
Due to the limited number of training samples per class, the tail-most class would only have one sample if the imbalance ratio is set to 100. 
Hence, we set the imbalance ratio to 10.
Fig.~\ref{fig:figure4} shows the number of training samples per class on different datasets.

\subsection{Experimental Setting}

\subsubsection{Network Architecture}
The spatial-temporal action exploration strategy and detached action-aware learning schedule of the proposed method are proposed for long-tailed skeleton-based action recognition, and can be combined with any action recognition backbones. 
Meanwhile, as mentioned before, various graph models do not have a large influence on the quality of representation learning. 
Hence, ST-GCN++~\cite{duan2022pyskl} is chosen as our backbone for the experiments. It is a recent representative approach using the GCN technique.

\subsubsection{Implementation Details}
In the experiments, the initial learning rate is set to 0.1 and decays with a cosine LR scheduler.
Moreover, stochastic gradient descent (SGD) optimization with the Nesterov momentum of 0.9 and the weight decay of $5\times10^{-4}$ is employed to tune the parameters.
Considering the large differences in the size of the datasets, we set the batch size for NTU 60, NTU 120, Northwestern-UCLA, and Kinetics Skeleton 400 datasets to 128, 128, 16, and 32, respectively.
The model is implemented with PyTorch deep learning framework.
And the maximum number of training epochs is set to 120.
We set $k=3$ using the grid search.
When adopting the detached action-aware learning schedule, the last 20 epochs are set as the detached stage. 
We set $\beta_{y} \in [0.99, 0.9999]$, i.e. $\upsilon=0.99$ and $\lambda=0.0099$.
As for the data pre-processing method and hyper-parameter settings, we follow the strategy in previous work~\cite{duan2022pyskl}.
The comparison methods use the same training and test sets as the proposed method in this paper, and these methods are implemented using official open-source code.
The same data augmentation is used for fair performance comparison.

\subsection{Experimental Results}

\subsubsection{Results of NTU RGB+D 60 dataset}
The proposed method is compared with quite a few SOTA methods on both X-sub and X-view benchmarks of the NTU 60 dataset. 
The comparisons under standard and long-tailed settings are displayed in Table~\ref{tab:NTU60}.
There are several important observations as follows.

In the long-tailed setting, there is a general degradation in the performance of previous methods. It verifies that current action recognition methods trained on balanced action datasets can not adapt well to long-tailed data.
Rather, the proposed method significantly improves the performance in the long-tailed setting, i.e., 81.8\% for X-sub benchmark and 85.4\% for X-view benchmark. The proposed method outperforms FR Head~\cite{zhou2023learning} by 4.5\% and 4.9\%, and the baseline method ST-GCN++ by 6.8\% and 5.4\% on the two benchmarks respectively.
In the standard setting, the proposed method also achieves the performance on par with other SOTA methods. The superiority is achieved by balanced representation learning, which generates more valuable skeletal samples and mitigates representation bias.

Note that many action recognition methods employ the multi-stream fusion framework to improve performance.
For a fair comparison, we follow the rule of these methods by fusing the results of four modalities.
The ensemble of 4-stream which includes joint, bone, joint motion, and bone motion is denoted as 4 ensemble. 
Here, motion means joint movement between two subsequent time frames.
In addition, we add the skip-modal and skip-motion-modal representations, so the proposed method can be denoted as 6 ensemble.
In Table~\ref{tab:NTU60}, the results of 4 ensemble and 6 ensemble are both reported in detail. In other tables, only the results of 6 ensemble are reported by default.

It is known that ST-GCN~\cite{yan2018spatial} is currently the most popular backbone model for skeleton-based action recognition.
Our backbone network, i.e. ST-GCN++~\cite{duan2022pyskl}, is actually an upgradation over ST-GCN, achieving better performance under the standard setting. 
However, under the long-tailed setting, the performances of the two models are relatively close.
The imbalanced data leads to skewed sample space and results in representation bias, ultimately degrading the representations learned by previous methods.
Integrated with the proposed method, the performance of ST-GCN++ under the long-tailed settings catches up with the accuracy of the original ST-GCN under the standard settings.
The proposed method generates more valuable samples to reshape the sample space, and mitigates existential bias in the representation space, effectively addressing the dilemmas of skewed sample space and representation bias.
The results imply that the proposed method can learn unbiased balanced representations from imbalanced data and handle imbalanced skeleton data well.

\subsubsection{Results of NTU RGB+D 120 dataset}
On the NTU 120 dataset, as shown in Table~\ref{tab:NTU120}, the proposed method also boosts the performance significantly when compared to other methods.
The proposed method achieves the accuracy of 69.7\% for the X-sub benchmark and 71.3\% for the X-set benchmark under the long-tailed setting.
In addition, the proposed method outperforms FR Head by 4.1\% on X-sub benchmark and 3.1\% on X-set benchmark, which is also a remarkable improvement.
Similar to the results on NTU 60 dataset, the performances demonstrate the effectiveness of the proposed method.

\subsubsection{Results of Northwestern-UCLA and Kinetics dataset}
For Northwestern-UCLA and Kinetics, Table~\ref{tab:nw_ucla} and Table~\ref{tab:Kinetics} tabulate the experimental results.
From Table~\ref{tab:nw_ucla}, it can be easily found that an obvious performance improvement has been attained by the proposed method.
The proposed method achieves an accuracy of 89.2\% for Northwestern-UCLA under the long-tailed setting, which outperforms InfoGCN~\cite{chi2022infogcn} by 17.2\%.
This improvement could be attributed to the relatively smaller sample size of Northwestern-UCLA.
The proposed method can generate more distinctive samples when the data is small, which can further enhance the diversity of the data and improve the model performance more effectively.
As shown in Table~\ref{tab:Kinetics}, it can be seen that the proposed method outperforms the performance of other SOTA methods by 0.9\% in Top-1 accuracy on  Kinetics Skeleton 400 dataset, which proves that our approach works well in the real-world long-tailed datasets.

\begin{table}[t]
\begin{center}
\caption{ Comparisons with SOTA methods on NTU 60 dataset in terms of classification accuracy (\%). $\dagger$~denotes ST-GCN++ as the baseline.}
\renewcommand\arraystretch{1.2}
{
\resizebox{\linewidth}{!}{
\begin{tabular}{l c|c c|c c}
\toprule
\multirow{2}{*}{Model} & \multirow{2}{*}{Published} 
& \multicolumn{2}{c|}{Standard} 
& \multicolumn{2}{c} {Long-tailed}\cr
& & X-sub &  X-view & X-sub &  X-view \cr
\midrule
ST-GCN \cite{yan2018spatial} & AAAI18 & 81.5 &  88.3 & 73.3 & 79.4 \\
2s-AGCN \cite{shi2019two} & CVPR19 & 88.5 & 95.1 & 74.0 & 78.9 \\
Shift-GCN \cite{cheng2020skeleton} & CVPR20 & 90.7 &  96.5 & 73.6 & 79.3 \\
MS-G3D \cite{liu2020disentangling} & CVPR20 & 91.5 & 96.2 & 74.8 & 80.9 \\
DC-GCN+ADG \cite{cheng2020decoupling} & ECCV20 & 90.8 &  96.6 & 75.0 & 79.7 \\
MST-GCN \cite{chen2021multi} & AAAI21 & 91.5 & 96.6 & 75.9 & 80.3 \\
CTR-GCN \cite{chen2021channel} & ICCV21 & 92.4 & 96.8 & 74.1  & 80.4 \\
EfficientGCN-B4 \cite{song2022constructing} & TPAMI22 & 91.7 & 95.7 & 75.2 & 80.9 \\
ST-GCN++$\dagger$ \cite{duan2022pyskl} & ACMMM22 & 92.6 & 97.4 & 75.0 & 80.0 \\
InfoGCN \cite{chi2022infogcn} & CVPR22 & 93.0 & 97.1 & 76.8 & 79.2 \\
FR Head~\cite{zhou2023learning} & CVPR23 & 92.8 & 96.8 & 77.3 & 80.5 \\
\midrule
Ours (Joint Only) & -- & 90.3 & 96.3 & 76.7 & 81.4 \\
Ours (Joint + Bone) & -- & 92.0 & 97.0 & 79.6  & 84.0 \\
Ours (4 ensemble) & -- & 92.4 & 97.1 & 81.0  & 84.9 \\
Ours (6 ensemble) & -- & 92.7 & 97.3 & \textbf{81.8}  & \textbf{85.4} \\
\bottomrule
\end{tabular}}
}
\label{tab:NTU60}
\end{center}
\vspace{-0.2cm}
\end{table}

\begin{table}[t]
\begin{center}
\caption{ Comparisons with SOTA methods on NTU 120 dataset in terms of classification accuracy (\%). $\dagger$~denotes ST-GCN++ as the baseline.}
\renewcommand\arraystretch{1.2}
{
\resizebox{\linewidth}{!}{
\begin{threeparttable}
\begin{tabular}{l c|c c|c c}
\toprule
\multirow{2}{*}{Model} & \multirow{2}{*}{Published} 
& \multicolumn{2}{c|}{Standard} 
& \multicolumn{2}{c} {Long-tailed}\cr
& & X-sub & X-set & X-sub & X-set \cr
\midrule
ST-GCN \cite{yan2018spatial} & AAAI18 & 70.7 & 73.2 & 59.1 & 62.3 \\
2s-AGCN \cite{shi2019two}& CVPR19 & 82.5 & 84.2 & 61.1 & 63.2 \\
Shift-GCN \cite{cheng2020skeleton} & CVPR20 & 85.9 &  87.6 & 62.3 & 64.5 \\
MS-G3D \cite{liu2020disentangling} & CVPR20 & 86.9 & 88.4 & 62.7 & 64.6 \\
DC-GCN+ADG \cite{cheng2020decoupling} & ECCV20 & 86.5 & 88.1 & 63.4 & 66.2 \\
MST-GCN \cite{chen2021multi} & AAAI21 & 87.5 & 88.8 & 63.8 & 65.9 \\
CTR-GCN \cite{chen2021channel} & ICCV21 & 88.9 & 90.6 & 63.2 & 66.0 \\
EfficientGCN-B4 \cite{song2022constructing} & TPAMI22 & 88.3 & 89.1 & 62.5 & 66.4 \\
ST-GCN++$\dagger$  \cite{duan2022pyskl} & ACMMM22 & 88.6 & 90.8 & 62.6 & 64.0 \\
InfoGCN \cite{chi2022infogcn} & CVPR22 & 89.8 & 91.2 & 64.2 & 67.1 \\
FR Head~\cite{zhou2023learning} & CVPR23 & 89.5 & 90.9 & 65.6 & 68.2 \\
\midrule
Ours (6 ensemble) & -- & 88.5 & 90.8 & \textbf{69.7} & \textbf{71.3} \\
\bottomrule
\end{tabular}
\end{threeparttable}
}
}
\label{tab:NTU120}
\end{center}
\vspace{-0.2cm}
\end{table}

\subsubsection{Comparisons with Algorithms for Long-tailed Data Distribution}
To demonstrate the advantages of the proposed method over current algorithms for long-tailed data distribution, we have conducted extensive experiments to compare with the following competing algorithms:
(1)~Cross-entropy (CE) loss;
(2)~Focal loss~\cite{lin2017focal}, which is a common re-weighting loss function for imbalanced class samples; 
(3)~Weighted softmax loss~\cite{huang2016learning}, which directly uses label frequencies of training samples for loss re-weighting; 
(4)~Random oversampling (ROS)~\cite{van2007experimental}, which oversamples minority samples to balance the class distribution in the training data;
(5)~Mixup~\cite{zhang2017mixup}, which generates new training data by mixing two samples and their labels;
(6)~Remix~\cite{chou2020remix}, which oversamples minority classes by assigning higher weights to the minority labels when using Mixup;
(7)~CB loss~\cite{cui2019class}, which introduces the effective number to approximate the expected sample number of different classes; 
(8)~Label-Distribution-Aware Margin Loss with deferred re-weighting (LDAM-DRW)~\cite{cao2019learning}, which regularizes the minority classes by minimizing a margin-based generalization bound; 
(9)~Equalization loss~\cite{tan2020equalization}, which ignores the gradients for rare categories;
(10)~Balanced Softmax~\cite{ren2020balanced}, which is an elegant unbiased extension of Softmax.
(11)~GumbelCE~\cite{alexandridis2022long}, which develops a Gumbel optimized strategy.
(12)~KPS loss~\cite{li2022key}, which presents a key point sensitive loss to improve the generalization performance of the classification model.

For fair comparisons, the same data augmentation is applied to all the compared algorithms in the experiments.
Training with cross-entropy (CE) loss is regarded as the baseline method. 
If any algorithm utilizes a different loss function, the CE loss is substituted for the corresponding loss.
Experiments are conducted on the long-tailed setting of NTU 60 and NTU 120.
The performance is measured by classification accuracy using the joint data.
Following~\cite{liu2019large}, we also report the accuracy of three disjoint subsets: Many-shot classes (classes with more than 100 training samples), Medium-shot classes (classes with 20 to 100 samples), and Few-shot classes (classes under 20 samples).

The performance comparison is shown in Table~\ref{tab:long_NTU60}.
It can be observed that, compared to vanilla training with CE loss, the proposed method maintains performance for the many-shot classes and significantly improves recognition accuracy for the medium-shot and few-shot classes. 
Meanwhile, the proposed method achieves significant performance advantages compared to other long-tailed methods.
For the many-shot classes, the performance improvements of different long-tailed algorithms are relatively close, but for the medium-shot and few-shot classes, the proposed method is superior.
The superiority is achieved through balanced learning of consensus knowledge and specific patterns of tail classes, resulting in more effective constraints being imposed on different classes. 
Balanced representation learning helps the model to understand the action data effectively regardless of the head or tail category, thus improving the overall recognition performance.

\begin{table*}[t]
\begin{center}
\caption{ 
Comparisons with SOTA long-tailed methods on the X-sub benchmark of NTU 60 (left) and NTU 120 (right).
Results measured by classification accuracy (\%) using the same backbone architecture.}
\renewcommand\arraystretch{1.2}
{
\setlength{\tabcolsep}{4.5mm}{
\begin{tabular}{l|c c c c|c c c c}
\toprule
\multirow{2}{*}{Method} & \multicolumn{4}{c|}{NTU 60 X-sub}
& \multicolumn{4}{c}{NTU 120 X-sub}\cr
& Overall & Many & Medium & Few & Overall & Many & Medium & Few\cr
\midrule
CE loss & 72.3 & 82.9 & 66.8 & 60.2 & 58.4 & 81.4 & 58.9 & 40.3 \\
Focal loss \cite{lin2017focal} & 72.4 & 82.7 & 67.1 & 61.8 & 58.5 & 80.9 & 59.1 & 45.2 \\
Weighted loss \cite{huang2016learning} & 72.6 & 83.1 & 68.1 & 58.8 & 58.6 & 81.9 & 58.2 & 42.1 \\
ROS \cite{van2007experimental} & 73.4 & 81.7 & 67.2 & 66.0 & 58.2 & 81.5 & 57.6 & 45.9 \\
Mixup \cite{zhang2017mixup} & 71.8 & 82.9 & 66.7 & 58.7 & 58.9 & 82.6 & 57.8 & 41.3 \\
Remix \cite{chou2020remix} & 71.7 & 82.2 & 66.3 & 62.2 & 58.4 & 81.8 & 56.8 & 41.2 \\
CB loss \cite{cui2019class} & 73.8 & 82.0 & 69.4 & 62.8 & 61.4 & 81.3 & 60.7 & 51.0 \\
LDAM-DRW \cite{cao2019learning} & 71.2 & 75.4 & 68.6 & 65.3 & 60.9 & 76.3 & 60.0 & 52.2 \\
Equalization loss \cite{tan2020equalization} & 73.7 & 84.7 & 69.3 & 58.5 & 59.4 & 82.8 & 58.8 & 47.5 \\
Balanced Softmax \cite{ren2020balanced} & 74.5 & 82.3 & 69.2 & 64.5 & 62.1 & \textbf{83.1} & 60.1 & 51.7 \\
GumbelCE \cite{alexandridis2022long} & 73.3 & 84.3 & 68.9 & 58.1 & 61.6 & 81.0 & 59.8 & 52.4 \\
KPS loss \cite{li2022key} & 73.6 & 83.1 & 72.3 & 61.6 & 62.4 & 82.5 & 61.2 & 52.1 \\
\midrule
Ours & \textbf{76.7} & \textbf{85.5} & \textbf{73.7} & \textbf{67.8} & \textbf{65.3} & 83.0 & \textbf{62.6} & \textbf{54.3} \\
\bottomrule
\end{tabular}
}
}
\label{tab:long_NTU60}
\end{center}
\vspace{-0.2cm}
\end{table*}

\subsection{Ablation Study}

In this subsection, we conduct ablation experiments and provide further investigations of the proposed method in four aspects: 
(1) What is the impact of different components?
(2) Can the proposed method be effectively integrated into other action recognition algorithms?
(3) Are the learned representations indeed superior to those learned by the original methods?
(4) How does the proposed method perform in the head, middle, and tail classes?
The experiments are launched on the long-tailed setting of NTU 60 dataset.

\subsubsection{Impact of Each Component}

To further analyze the effectiveness of the proposed method, we conducted additional experiments to validate the impact of the spatial-temporal action exploration strategy and the detached action-aware learning schedule. 
The performance is measured by classification accuracy on NTU 60 using the joint data.

\begin{table}[t]
\begin{center}
\caption{Comparisons with SOTA methods on the Northwestern-UCLA dataset in terms of classification accuracy (\%). }
\renewcommand\arraystretch{1.2}
{
\setlength{\tabcolsep}{9mm}{
\begin{threeparttable}
\begin{tabular}{l|c}
\toprule
\multirow{2}{*}{Model}  
& {Long-tailed} \cr
& Top1 \cr
\midrule
2s-AGCN \cite{shi2019two} & 70.1 \\
MS-G3D \cite{liu2020disentangling} & 72.8 \\
CTR-GCN \cite{chen2021channel} & 71.6 \\
InfoGCN \cite{chi2022infogcn} & 72.0 \\
\midrule
Ours (6 ensemble) & \textbf{89.2} \\
\bottomrule
\end{tabular}
\end{threeparttable}
}
}
\label{tab:nw_ucla}
\end{center}
\vspace{-0.2cm}
\end{table}

\begin{table}[tp]
\begin{center}
\caption{Comparisons with SOTA methods on the
Kinetics Skeleton 400 dataset in terms of classification accuracy (\%). }
\renewcommand\arraystretch{1.2}
{
\setlength{\tabcolsep}{6mm}{
\begin{threeparttable}
\begin{tabular}{l | c}
\toprule
Model & Top1  \cr
\midrule
TCN \cite{kim2017interpretable} & 20.3 \\
ST-GCN \cite{yan2018spatial} & 30.7 \\
AS-GCN \cite{li2019actional} & 34.8 \\
2s-AGCN \cite{shi2019two} & 36.1 \\
MS-G3D \cite{liu2020disentangling} & 38.0 \\
PoseConv3D \cite{duan2022revisiting} & 47.7 \\
\midrule
Ours (6 ensemble) & \textbf{48.6} \\
\bottomrule
\end{tabular}
\end{threeparttable}
}
}
\label{tab:Kinetics}
\end{center}
\vspace{-0.2cm}
\end{table}

Firstly, we analyze the spatial-temporal action exploration (STAE) and compare the accuracy gains of rebalanced partial mixup (RPM) and temporal reverse perception (TRP) in Table~\ref{tab:aug}.
The STAE strategy leads to significant improvements, with gains of 3.4\% on X-sub benchmark and 2.9\% on X-view benchmark. 
It is proven by the results that both RPM and TRP contribute to the action recognition model.
The STAE strategy strives to aggregate spatial structure information and make full use of temporal distinctive information.
The experimental results demonstrate that the proposed strategy generates more meaningful skeletal samples, and the valuable data further helps the model to understand the long-tailed action.

\begin{figure*}[ht]
\vspace{-0.2cm}
\centering
\subcaptionbox{\label{fig: 5a}}
{\includegraphics[width=0.24\linewidth]{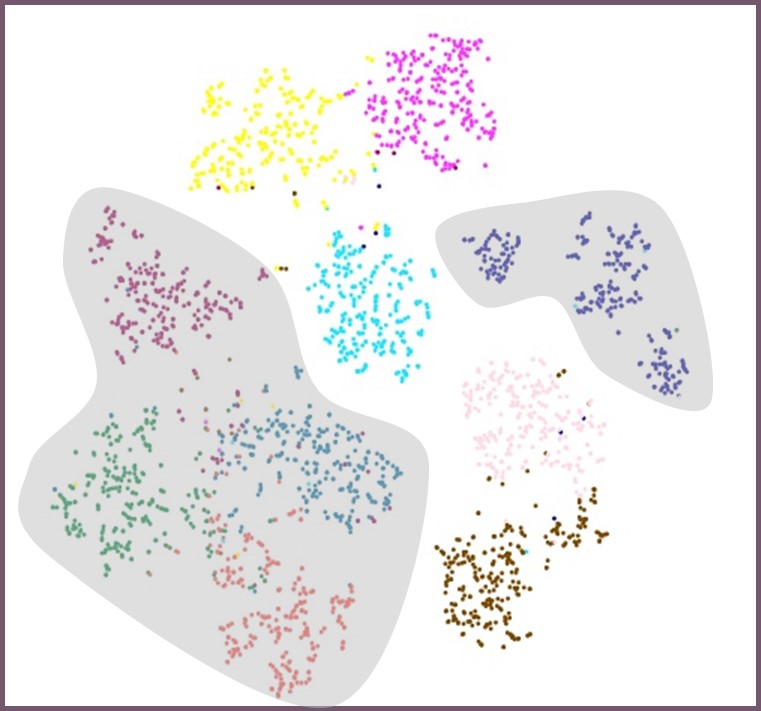}}
\subcaptionbox{\label{fig: 5b}}
{\includegraphics[width=0.24\linewidth]{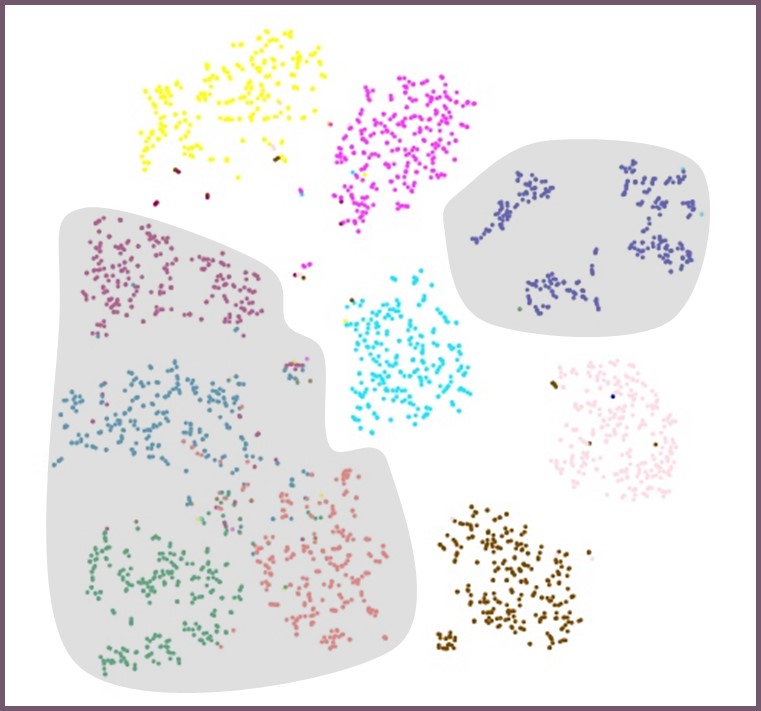}}
\subcaptionbox{\label{fig: 5c}}
{\includegraphics[width=0.24\linewidth]{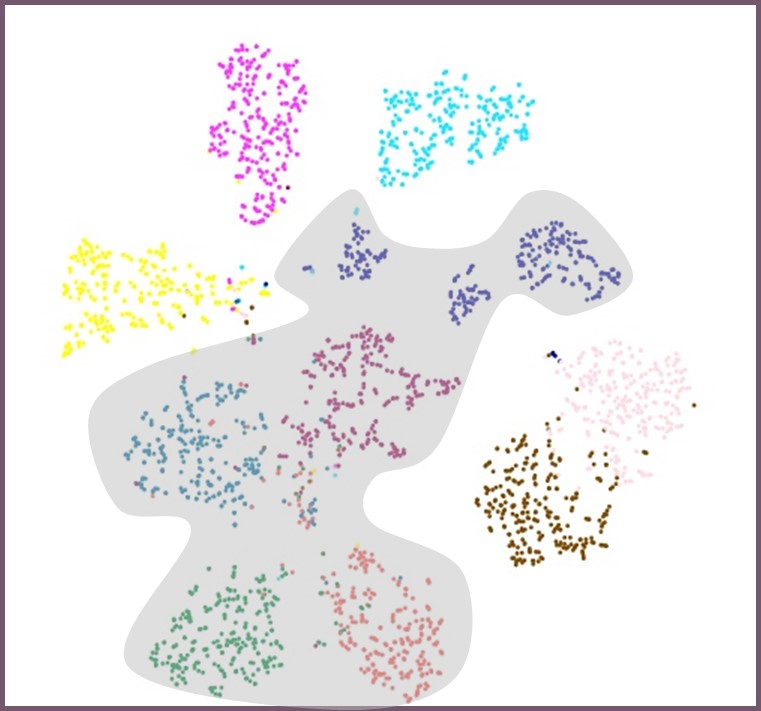}}
\subcaptionbox{\label{fig: 5d}}
{\includegraphics[width=0.24\linewidth]{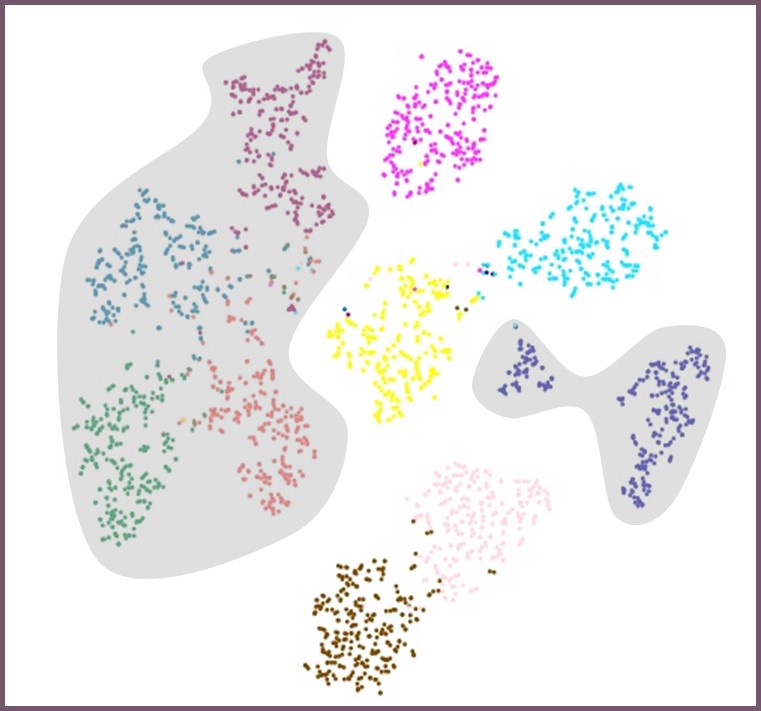}}
\caption{ 
The \textbf{t-SNE} visualization of the features learned by (a) Baseline, (b) Baseline with augmentation, (c) The proposed method with only the STAE strategy, and (d) The proposed balanced representation learning method.
Different colors indicate different classes.
Note that the learned representations of five head classes and five tail classes are visualized.
The sample points of tail classes are overlaid by grey area to show the improvements brought by different components.
}
\label{fig:figure5}
\vspace{-0.2cm}
\end{figure*}

\begin{table}[t]
\begin{center}
\caption{ 
Comparisons of classification accuracies (\%) based on (top) removing RPM or TRP from the STAE strategy and (bottom) different learning schedules.
}
\renewcommand\arraystretch{1.2}
{
\centering
\setlength{\tabcolsep}{4mm}{
\begin{tabular}{ l | c c }
\toprule
Model Configurations & X-sub & X-view \\
\midrule
\midrule
Baseline (CE loss) & 72.3 & 76.4 \\
\ \, w/ RPM & 74.8 & 78.5 \\
\ \, w/ TRP & 74.9 & 78.1 \\
\ \, w/ STAE ( RPM \& TRP) & \textbf{75.7} & \textbf{79.3} \\
\midrule
\midrule
CE loss & 75.7 & 79.3 \\
Focal loss & 75.3 & 78.9 \\
Weighted loss & 71.0 & 76.1 \\
Action-Aware loss & 75.6 & 80.6 \\
\midrule
Detached Focal loss & 75.9 & 79.8 \\
Detached Weighted loss & 74.3 & 78.5 \\
Detached Action-Aware loss & \textbf{76.7} & \textbf{81.4} \\
\bottomrule
\end{tabular}
}
}
\label{tab:aug}
\end{center}
\vspace{-0.2cm}
\end{table}

Secondly, to validate the efficacy of the proposed detached action-aware learning schedule, we construct different ablation experiments and report their performance in Table~\ref{tab:aug}. 
It can be seen that directly using re-weighting losses throughout the training process leads to a certain degree of performance degradation.
Despite the performance of focal loss being relatively stable, the gap between the weighted loss and CE loss is obvious.
In addition, for the X-sub benchmark, the accuracy of the action-aware loss also drops by 0.1\%. 
And then the detached learning schedule is adopted. 
It can be observed that the performances of different losses become generally better. 
Meanwhile, the detached action-aware loss is more stable, bringing a performance boost compared with the original CE loss.

Finally, we scrutinize the contribution of each component.
The results are listed in Table~\ref{tab:ablation}.
$\rm C_{STAE}$ and $\rm C_{DAA}$ refer to the spatial-temporal action exploration strategy and detached action-aware learning schedule respectively, and $\rm C_{MS}$ denotes the ensemble of multiple streams.
Accordingly, the absolute accuracy and $\Delta$ are reported.
$\Delta$ represents the relative accuracy improvement over the baseline ST-GCN++~\cite{duan2022pyskl} by using combinations of different components.
As shown in Table~\ref{tab:ablation}, we can observe that the performance of the model is improved by different components.
By comparing the performance of variants denoted by $\mathcal{M}_{1}$, $\mathcal{M}_{2}$, $\mathcal{M}_{3}$, and $\mathcal{M}_{4}$, $\rm C_{STAE}$, $\rm C_{DAA}$, and $\rm C_{MS}$ bring 3.4\%, 2.6\%, and 0.8\% improvement respectively. 
By mining multiple streams, the model can obtain more valuable information for action recognition. 
After combining all components, the method results in $\Delta=6.8\%$ improvement on the NTU-60 X-sub setting.
The results indicate that the components of the proposed method contribute to understanding the long-tailed action data.

\begin{table}[tp]
\begin{center}
\caption{ 
Comparisons with different components on X-sub benchmark in terms of classification accuracy (\%). 
$\Delta$ represents the relative accuracy improvement over the baseline, ST-GCN++~\cite{duan2022pyskl}.
}
\renewcommand\arraystretch{1.2}
{
\resizebox{\linewidth}{!}{
\begin{tabular}{ l c c c c c c}
\toprule
Method & ST-GCN++ & $\rm C_{STAE}$ & $\rm C_{DAA}$ & $\rm C_{MS}$ & Accuracy & $\Delta$ \\
\midrule
$\mathcal{M}_{1}$ & \checkmark &  &  &  & 75.0 & -- \\
$\mathcal{M}_{2}$ & \checkmark & \checkmark &  &  & 78.4 & $\uparrow$ 3.4 \\
$\mathcal{M}_{3}$ & \checkmark & \checkmark & \checkmark &  & 81.0 & $\uparrow$ 6.0 \\
$\mathcal{M}_{4}$ & \checkmark & \checkmark & \checkmark & \checkmark & \textbf{81.8} & $\uparrow$ 6.8 \\
\bottomrule
\end{tabular}
}
}
\label{tab:ablation}
\end{center}
\vspace{-0.2cm}
\end{table}

\begin{table}[tp]
\begin{center}
\caption{ Comparisons with various backbones on the NTU 60 dataset in terms of classification accuracy (\%). 
}
\renewcommand\arraystretch{1.2}
{
\setlength{\tabcolsep}{3.5mm}{
\begin{threeparttable}
\begin{tabular}{l|cc|cc}
\toprule
\multirow{2}{*}{Model} & \multicolumn{2}{c|}{X-sub} & \multicolumn{2}{c}{X-view} \cr
& Vanilla & +BRL & Vanilla & +BRL \cr
\midrule
ST-GCN++ \cite{duan2022pyskl} & 75.0 & 81.8 & 80.0 & 85.4 \cr 
& +(0.0) & \textbf{(+6.8)} & +(0.0) & \textbf{(+5.4)} \\
MS-G3D \cite{liu2020disentangling} & 74.8 & 82.4 & 80.9 & 85.7 \cr
& +(0.0) & \textbf{(+7.6)} & +(0.0) & \textbf{(+4.8)} \\
CTR-GCN \cite{chen2021channel} & 74.1 & 81.8 & 80.4 &  85.0 \cr
& +(0.0) & \textbf{(+7.7)} & +(0.0) & \textbf{(+4.6)} \\
\bottomrule
\end{tabular}
\end{threeparttable}
}
}
\label{tab:combination}
\end{center}
\vspace{-0.2cm}
\end{table}

\subsubsection{Integration With Other Algorithms}

The proposed balanced representation learning method~(BRL) also has a superior interoperability capacity and is easy to incorporate into other methods.
In this subsection, the scalability of the proposed method is assessed on NTU 60 with a combination of various skeleton-based action recognition backbones. 
We apply the proposed BRL to popular recognition methods, i.e. ST-GCN++~\cite{duan2022pyskl}, MS-G3D~\cite{liu2020disentangling} and CTR-GCN~\cite{chen2021channel}.

The results with different recognition backbones are tabulated in Table~\ref{tab:combination}.
For ST-GCN++, the method results in a 6.8\% gain on X-sub benchmark and a 5.4\% gain on X-view benchmark. 
Combined with the proposed method, the performances of other algorithms are also improved to some extent.
For example, for MS-G3D, the method brings as much as 7.6\% gain on X-sub benchmark and 4.8\% gain on X-view benchmark.
And for CTR-GCN, the method brings a 7.7\% gain on X-sub benchmark and a 4.6\% gain on X-view benchmark.
The results show that our method consistently improves the performance of different action recognition backbones.
It is inspiring that the proposed method can be easily integrated into various action recognition algorithms.

\begin{figure*}[t]
\centering
\subcaptionbox{Per-group error rates on NTU 60-LT\label{fig: 6a}}
{\includegraphics[width=0.49\linewidth]{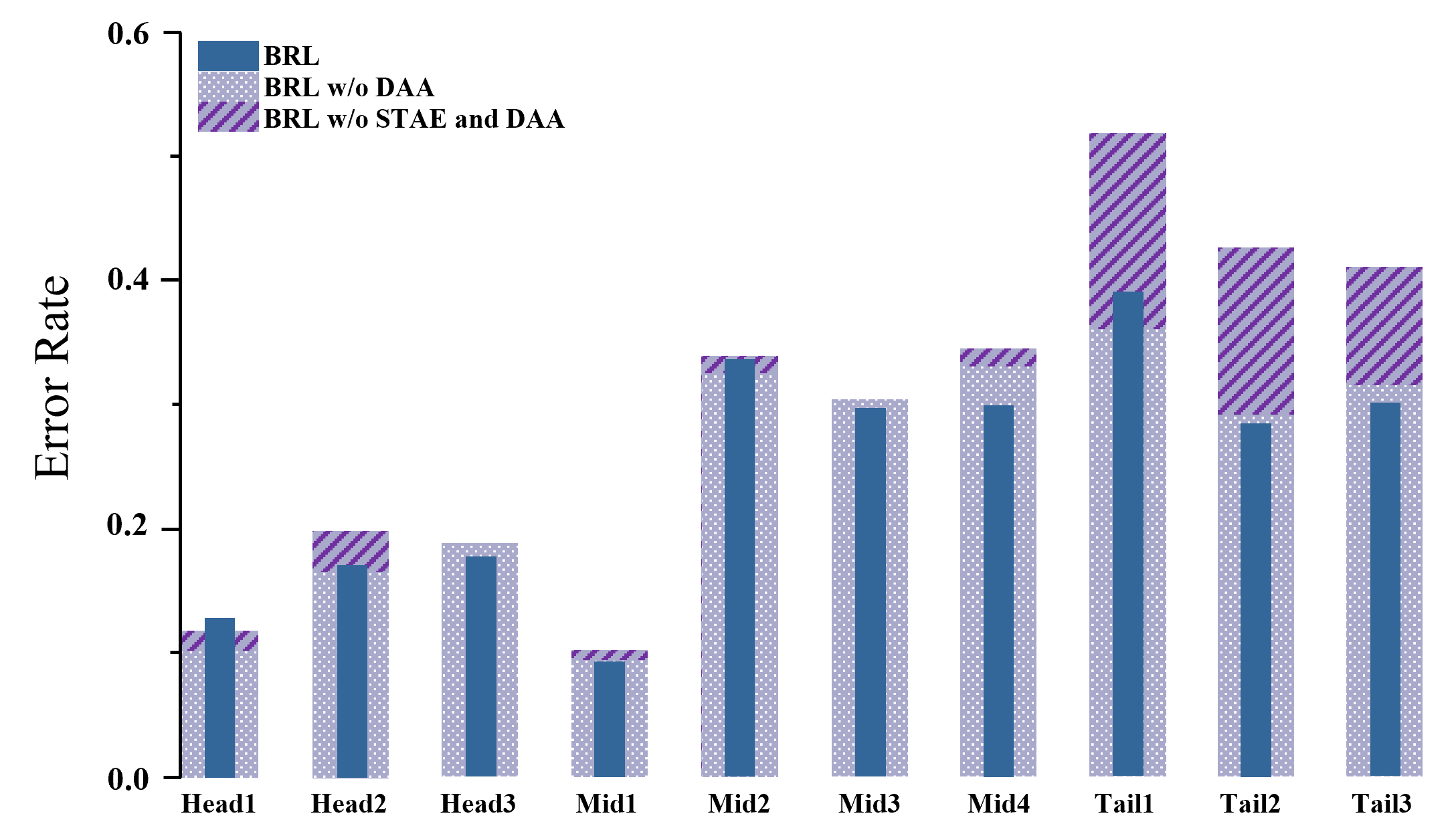}}
\subcaptionbox{Per-group error rates on NTU 120-LT\label{fig: 6b}}
{\includegraphics[width=0.49\linewidth]{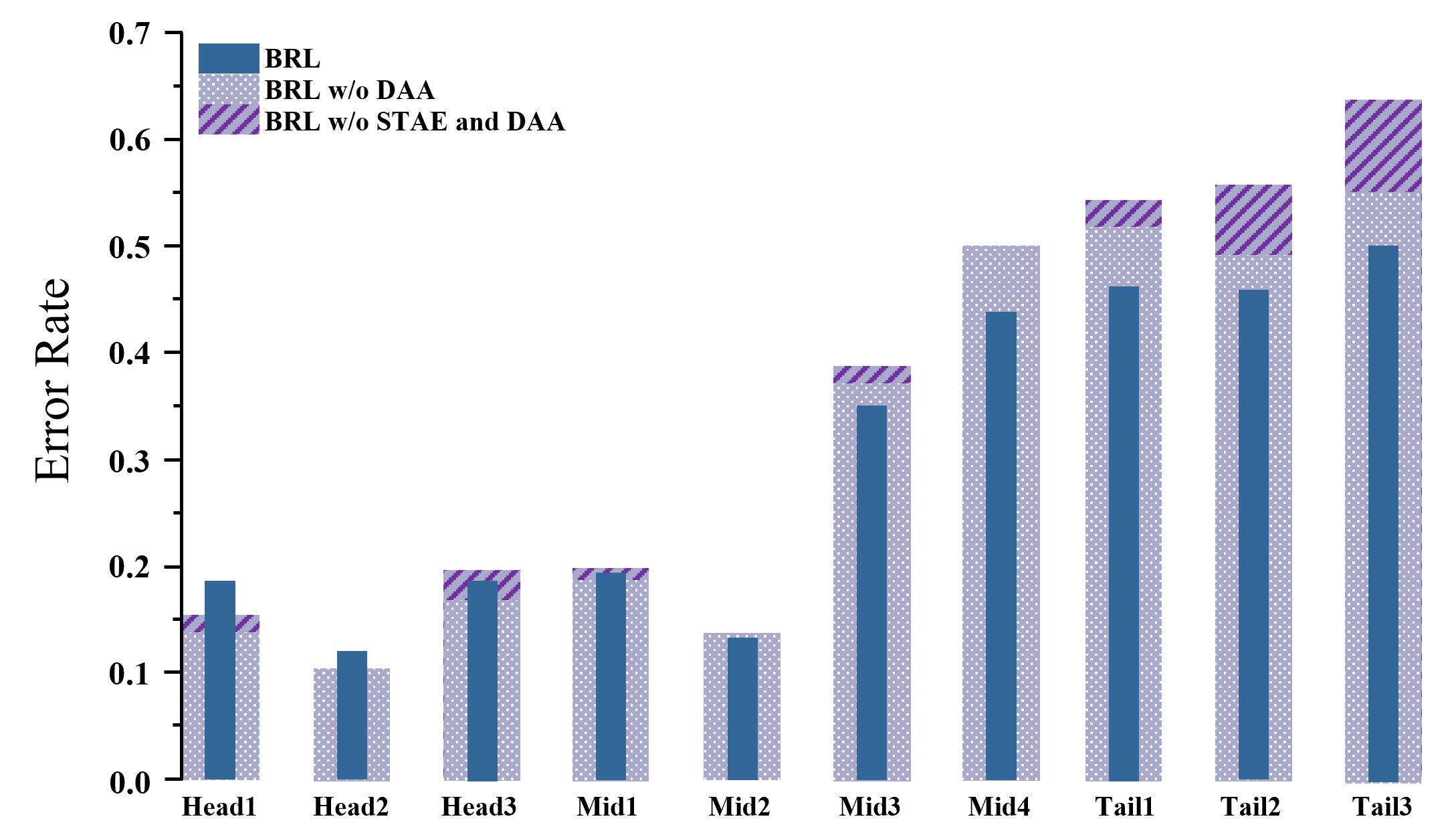}}
\caption{
Error rates obtained for different combinations of components on the NTU 60-LT/NTU 120-LT dataset.
For NTU 60-LT/NTU 120-LT, the classes are aggregated into 10 groups.
Head classes are with low indices. Conversely, tail classes are with higher indices. 
Lower error rates indicate better performance.
}
\label{fig:figure6}
\end{figure*}

\begin{figure*}[t]
\centering
\subcaptionbox{Head, middle and tail error rates on NTU 60-LT\label{fig: 7a}}
{\includegraphics[width=0.49\linewidth]{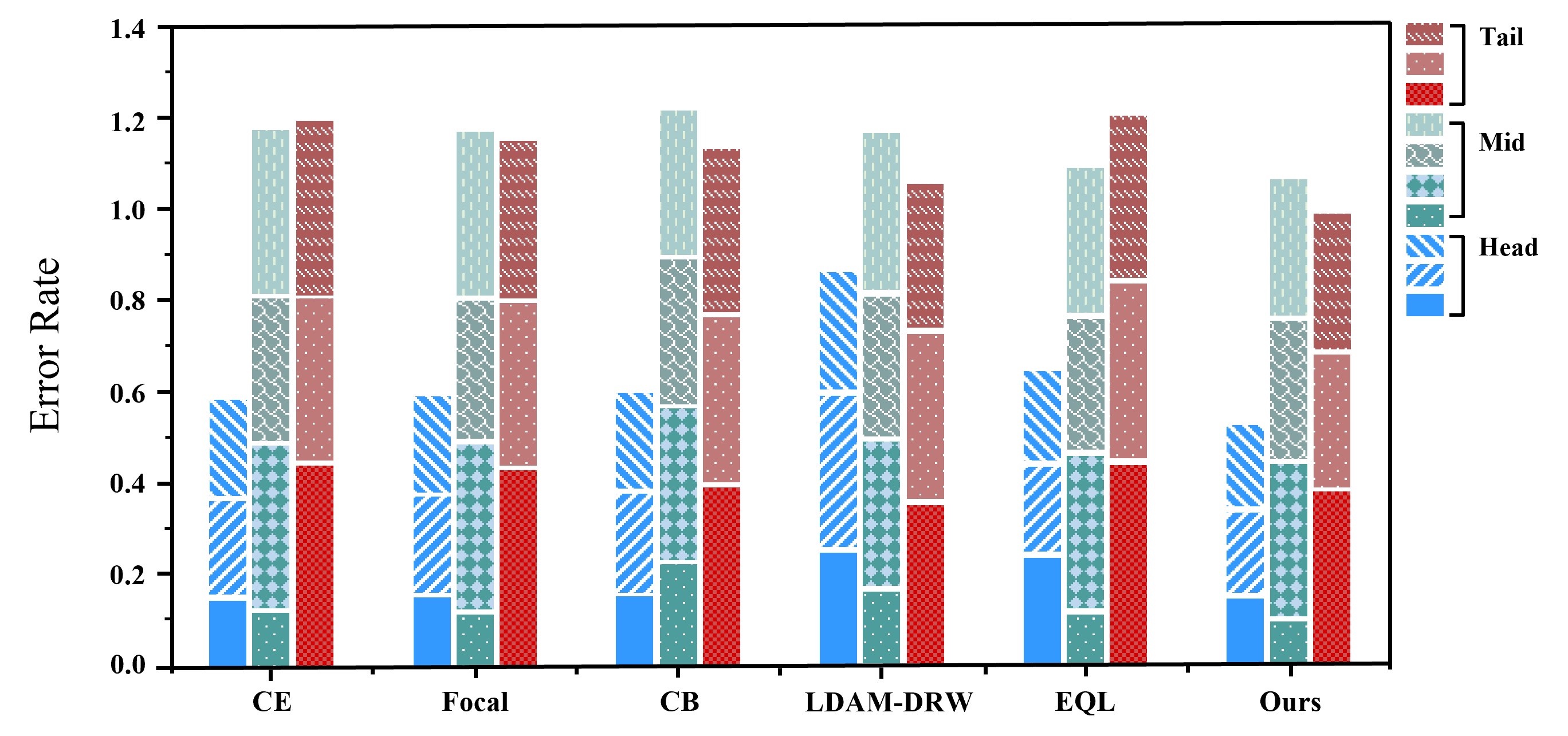}}
\subcaptionbox{Head, middle and tail error rates on NTU 120-LT\label{fig: 7b}}
{\includegraphics[width=0.49\linewidth]{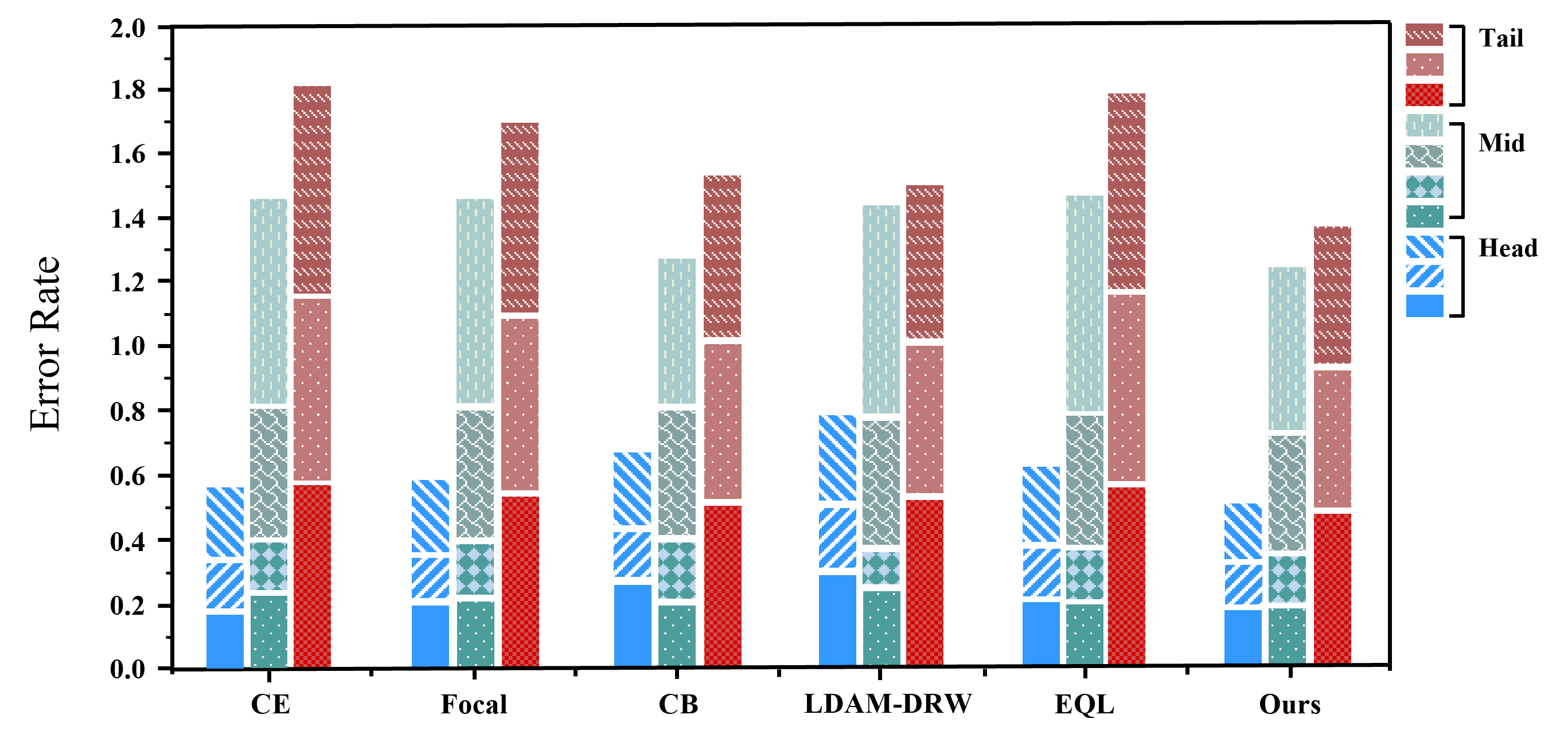}}
\caption{
Error rates obtained by different long-tailed algorithms on the NTU 60-LT/NTU 120-LT dataset.
For easy comparisons, the results are further integrated into the head, middle, and tail parts.
Lower error rates indicate better performance.
}
\label{fig:figure7}
\end{figure*}

\subsubsection{Visualization of Learned Representations}

To further verify the effectiveness of the proposed method, the learned representations of five head categories and five tail categories are visualized in Fig.~\ref{fig:figure5}.
It exhibits the t-SNE embeddings of the latent representations across different methods. 
The sample points of tail classes are overlaid by grey area to show the improvements brought by different components. 
It can be observed that the proposed method can obtain more accurate classification boundaries for tail classes.
With more valuable data and more effective weighting constraints, the proposed method can indeed learn balanced representations for long-tailed skeleton action datasets.

\subsubsection{Improvement on the Head, Middle, and Tail Classes}

To analyze the performance of the proposed method on different action classes, we construct the following experiments.

On the one hand, based on Table~\ref{tab:ablation}, we further analyze the effects of different components on the performance of different classes.
As shown in Fig.~\ref{fig:figure6}, three schemes are tested on the X-sub benchmark of NTU 60-LT and NTU 120-LT, including BRL, BRL without DAA, and BRL without STAE and DAA.
In order to facilitate the visualization, we aggregate the classes into 10 groups according to the label frequency.
To be specific, for NTU 60, the dataset has 60 classes, and each group includes 6 classes.
The most frequent 3 groups are named Head1 to Head3. 
The 4 groups with medium frequency are named Mid1 to Mid4, and the rest 3 least frequent groups are named Tail1 to Tail3 in order. 
Per-class error rates on the NTU 60-LT/NTU 120-LT datasets are plotted.
The results verify that the proposed components effectively improve the accuracy of tail classes and that the proposed method can simultaneously maintain the performance of head classes well.

On the other hand, we compare the proposed method with other related methods, such as Focal loss~\cite{lin2017focal}, CB loss~\cite{cui2019class}, LDAM-DRW~\cite{cao2019learning}, and Equalization loss~\cite{tan2020equalization}.
The experiments are conducted on the X-sub benchmark of NTU 60-LT and NTU 120-LT.
We also aggregate the classes into ten groups based on the label frequency and further integrate the results into the head, middle, and tail parts for ease of comparison.
The experimental results presented in Fig.~\ref{fig:figure7} reveal a common trend across different methods, where performance is better for head classes and worse for middle and tail classes. 
This can be attributed to the fact that the head class is typically more diverse in terms of available samples, resulting in more consistent recognition performance. 
Additionally, compared to the baseline CE loss, other methods such as Focal loss, CB loss, and LDAM-DRW show improvements in reducing the error rates for middle and tail classes, while EQL is relatively less effective. 
Finally, in comparison with other long-tailed methods, the proposed method can learn balanced representations of the tail classes more effectively, which leads to a significant reduction in the error rates of tail classes. The proposed method can maintain the performance of head classes as well.

\section{Conclusion}\label{paper_5}

In this work, we make two major contributions to address the long-tailed dilemmas in skeleton-based action recognition: the spatial-temporal action exploration strategy to generate more valuable samples, and the detached action-aware learning schedule to mitigate representation bias. 
In addition, we further introduce a skip-modal representation of the human skeleton for model ensemble. 
By coupling these components, the balanced representation learning method is proposed. 
Extensive experiments are conducted on several long-tailed simulated and real-world datasets.
The imbalanced data actually degrades the representations learned by current methods. 
Experimental results show that the proposed method learns unbiased balanced representations from imbalanced action data and outperforms state-of-the-art methods.
We believe our method can aid the development of skeleton-based action recognition, and provide a new perspective to understanding long-tailed action data.

\bibliographystyle{IEEEtran}
\bibliography{trans}

\end{document}